\documentclass[british,english]{article}
\usepackage[latin9]{inputenc}
\usepackage{color}
\usepackage{babel}
\usepackage{float}
\usepackage{url}
\usepackage{bm}
\usepackage{amsmath}
\usepackage{graphicx}
\usepackage[numbers]{natbib}
\usepackage[unicode=true,
 bookmarks=true,bookmarksnumbered=false,bookmarksopen=false,
 breaklinks=false,pdfborder={0 0 0},backref=false,colorlinks=true]
 {hyperref}
\hypersetup{pdftitle={A polygon-based interpolation operator for super-resolution imaging},
 pdfauthor={Stefan J. van der Walt and B. M. Herbst},
 pdfkeywords={super-resolution, image processing, least squares},
 linkcolor=blueish,citecolor=blueish,filecolor=blueish,urlcolor=blueish}

\makeatletter

\newcommand{\lyxdot}{.}

\floatstyle{ruled}
\newfloat{algorithm}{tbp}{loa}
\providecommand{\algorithmname}{Algorithm}
\floatname{algorithm}{\protect\algorithmname}

\usepackage{nips12submit_e,times}
\nipsfinalcopy

\pdfoutput=1

\usepackage{natbib}

\definecolor{blueish}{rgb}{0.3,0.6,1}

\@ifundefined{showcaptionsetup}{}{%
 \PassOptionsToPackage{caption=false}{subfig}}
\usepackage{subfig}
\makeatother

\addto\captionsbritish{\renewcommand{\algorithmname}{Algorithm}}

\begin{document}

\title{A polygon-based interpolation operator for super-resolution imaging}

\author{Stéfan J. van der Walt\\
Applied Mathematics,\\
Stellenbosch University \\
\texttt{stefan@sun.ac.za} \And B. M. Herbst\\
Applied Mathematics\\
Stellenbosch University\\
\texttt{herbst@sun.ac.za}}
\maketitle
\begin{abstract}
We outline the super-resolution reconstruction problem posed as a
maximization of probability. We then introduce an interpolation method
based on polygonal pixel overlap, express it as a linear operator,
and use it to improve reconstruction. Polygon interpolation outperforms
the simpler bilinear interpolation operator and, unlike Gaussian modeling
of pixels, requires no parameter estimation. A free software implementation
that reproduces the results shown is provided.
\end{abstract}

\section{Introduction}

\selectlanguage{british}%
Super-resolution imaging is the process whereby several low-resolution
photographs of an object are combined to form a single high-resolution
estimation. The problem, as most commonly viewed today, was outlined
by Cheeseman \emph{et al.} \citep{Cheeseman1993}, and has as its
solution a maximum a posteriori point estimate. Given a number of
low-resolution images, concatenated to form the vector $\mathbf{b}$,
as well as accompanying camera alignment parameters, $\mathbf{c}$,
the super-resolution problem is the estimation of a high-resolution
image, $\mathbf{x}$, such that the posterior
\[
P(\mathbf{x}|\mathbf{b},\mathbf{c})
\]
is maximized. The problem becomes more tractable if viewed, via Bayes's
theorem, as the maximization of
\begin{equation}
P(\mathbf{x}|\mathbf{b},\mathbf{c})=\frac{P(\mathbf{b}|\mathbf{x},\mathbf{c})P(\mathbf{x}|\mathbf{c})}{P(\mathbf{b}|\mathbf{c})}.\label{eq:SR_Bayes}
\end{equation}
\foreignlanguage{english}{We can also think of it as the maximum likelihood
estimator, regularized by the term }$P(\mathbf{x}|\mathbf{c})$, which
answers the question: ``Given a high-resolution image of a scene
and camera positions, how would the obtained low-resolution images
look?''.\foreignlanguage{english}{ }

\selectlanguage{english}%
We compute the solution to the the above problem, after assuming that
a linear relationship, $A{\bf x}={\bf b}$, exists between the high-resolution
reconstruction $\mathbf{x}$ and the low-resolution input images $\mathbf{b}$.
We explore the structure of $A$, which incorporates aspects such
as geometric distortion and interpolation, and show that an interpolation
operator based on pixel overlap has desirable properties.

\section{Image formation model}

\selectlanguage{british}%
Image formation is the process whereby light, reflected from a scene,
travels through an optical system and causes accumulation of charge
in a photosensitive sensor element. The charge values are read out,
amplified, discretised and possibly processed before being stored
as image intensity values.

Super-resolution relies on slight shifts in camera (or object) positions
between several input frames to provide high-frequency information
lost during sampling. Images are registered, preferably to sub-pixel
accuracy, before one of several methods (averaging or ``stacking'',
map and deblur \citep{Guichard1999}, pan-sharpening \citep{Levin2004,Garzelli},
supervised super-resolution \citep{Wright2008}, etc.) is applied
to restore high-frequency detail.

The image acquisition process is often represented as the simplified
model
\begin{equation}
\mathbf{b}^{(i)}=S\downarrow(h(\mathcal{T}(\mathbf{x})))+\bm{\eta}\label{eq:acquisition}
\end{equation}
where $\mathbf{b}^{(i)}$ is the $i$-th low-resolution digital image
obtained, $\mathbf{x}$ is a high-resolution representation of scene
radiance, $\mathcal{T}$ is a geometric transformation dependent on
camera position, $h$ is the camera point-spread function, $S\downarrow$
is the downsampling operator and $\bm{\eta}$ is normally distributed
noise.

This model makes two main assumptions: that low-resolution images
can be reproduced from the high-resolution image, and that any additive
noise is linear, zero-mean and Gaussian. Since, in reality, the camera
generates low-resolution images based on the scene itself, the first
assumption holds only if the high-resolution image is a fairly good
representation of the true scene radiance. As for the second, noise
sources vary according to exposure level \citep{Faraji2006}, but
among the different prominent sources (readout, shot and fixed pattern
\citep{Janesick2001}) only fixed-pattern noise has non-zero mean
\citep{Healey1994}, and can be removed using flat-field correction
\citep{Fridrich2009} without adversely affecting reconstruction.

Estimating the high-resolution image from a set of low-resolution
images only becomes tractable once further assumptions are made. In
our case, we make the fairly weak assumption that the linear relationship
\[
\mathbf{b}^{(i)}=A^{(i)}\mathbf{x}+\bm{\eta}
\]
holds, i.e., that the intensity of a pixel in any of the low-resolution
images may be obtained as a linear combination of corresponding neighboring
pixels in the high-resolution image.

\selectlanguage{english}%

\section{The maximum likelihood solution}

\selectlanguage{british}%
Experiments show \citep{Capel2001,Cheeseman1993} that the per-pixel
intensity probability, $P(b|\mathbf{x},\mathbf{c})$ with $b\in\mathbf{b}$,
can be approximated as a Gaussian distribution,
\[
P(b|\mathbf{x},\mathbf{c})=\mathcal{N}(\bar{b},\sigma_{b})=\frac{1}{\sqrt{2\pi\sigma_{b}^{2}}}e^{-\left(b-\bar{b}\right){}^{2}/\left(2\sigma_{b}^{2}\right)},
\]
with the $\mathbf{x}$-dependence through $\mathbf{b}=A\mathbf{x}$.
If we assume independence of pixels across different low-resolution
images (reasonable, given that multiple factors such as noise and
registration errors are at play), as well as independence over neighboring
pixels (since neighborhood influence is highly localised), the distribution
becomes

\[
P(\mathbf{b}|\mathbf{x},\mathbf{c})=\frac{1}{\left|2\pi\Sigma\right|^{1/2}}\exp\left(-\frac{1}{2}(\mathbf{b}-A\mathbf{x})^{T}\Sigma^{-1}(\mathbf{b}-A\mathbf{x})\right),
\]
with $\bar{\mathbf{b}}$ set to $A\mathbf{x}$, according to our assumed
model. Maximizing the log probability under the assumptions of spherical
covariance $\Sigma=\sigma^{2}I$, and prior $P(\mathbf{x}|\mathbf{c})$
Gaussian with zero mean, yields

\begin{eqnarray}
\arg\max_{\mathbf{x}}P(\mathbf{b}|\mathbf{x},\mathbf{c})=\hat{\mathbf{x}} & = & \arg\min_{\mathbf{x}}\left[\left\Vert \mathbf{b}-A\mathbf{x}\right\Vert ^{2}+\lambda\mathbf{x^{T}}\mathbf{x}\right].\label{eq:minimise_regularise}
\end{eqnarray}
The form of \eqref{eq:minimise_regularise} is well known as the regularised
or damped solution to the least-squares problem 
\[
A\mathbf{x}=\mathbf{b}.
\]

Yet, the assumption of a zero-mean prior is known to be invalid---the
output image is unlikely to be black. We therefore transform the problem
to seek a solution around an arbitrary prior estimate, $\mathbf{x}_{0}$:\foreignlanguage{english}{
\begin{eqnarray}
\hat{\mathbf{b}} & = & \mathbf{b}-A\mathbf{x}_{0}\nonumber \\
\boldsymbol{\delta}\hat{\mathbf{x}} & = & \arg\min_{\boldsymbol{\delta}\mathbf{x}}\left\{ \left\Vert A\bm{\delta}\mathbf{x}-\mathbf{\hat{b}}\right\Vert _{2}^{2}+\lambda(\boldsymbol{\delta}\mathbf{x})^{T}(\boldsymbol{\delta}\mathbf{x})\right\} \label{eq:solve_delta}\\
\hat{\mathbf{x}} & = & \mathbf{x}_{0}+\boldsymbol{\delta}\hat{\mathbf{x}}.\label{eq:prior_plus_delta}
\end{eqnarray}
Here, the difference between the prior estimate, $\mathbf{x}_{0}$,
and the reconstructed image, $\hat{\mathbf{x}}$, is expected to be
distributed around zero.}

\section{The image formation matrix, $A$}

In this section, we discuss the structure of the image formation matrix
$A^{(i)}$, that approximates \eqref{eq:acquisition} by encapsulating
three processes: geometric transformation, the effect of the point-spread
function as well as down-sampling. More concretely, what do the values
in $A^{(i)}$ represent, and how can they be computed?

First, consider the dimensions of the vectors and matrices involved.
The noiseless model for a single image, $A^{(i)}\mathbf{x}=\mathbf{b}^{(i)}$,
includes the $i$-th low-resolution output image, $\mathbf{b}^{(i)}$,
a $P\times Q$ image unpacked into a vector. For notational convenience,
assign $M=PQ$. The vector $\mathbf{x}$ is the high-resolution $zP\times zQ$
image, also unpacked. The zoom factor, $z$ with $z>1$, represents
the increase in resolution; e.g., if $z=2$ then the high-resolution
image has twice as many pixels as the low-resolution image along each
axis. The dimensionality of \textbf{$\mathbf{x}$} is $N=z^{2}M$.
Given the dimensions of $\mathbf{b}^{(i)}$ and\textbf{ $\mathbf{x}$},
the shape of the matrix $A$ has to be $M\times N=M\times z^{2}M$.

Since $M<N$, the system $A^{(0)}\mathbf{x}=\mathbf{b}^{(0)}$ is
underdetermined, but when combining all $k$ transformation-interpolation
matrices,
\[
A=\left[\begin{array}{c}
A^{(0)}\\
A^{(1)}\\
\vdots\\
A^{(k)}
\end{array}\right]\quad\mbox{and}\quad\mathbf{b}=\left[\begin{array}{c}
\mathbf{b}^{(0)}\\
\mathbf{b}^{(1)}\\
\vdots\\
\mathbf{b}^{(k)}
\end{array}\right]
\]
the resulting system $A\mathbf{x}=\mathbf{b}$ is overdetermined as
long as $k>z^{2}$. Note that, even when combining a large number
of images, there is no guarantee that each additional image provides
independent information (for example, the case in which $k$ identical
frames are combined).

Returning to $A^{(i)}$, each row represents weights for values in
$\mathbf{x}$, combined to form a single pixel of $\mathbf{b}^{(i)}$,
such that the $m$-th pixel is
\begin{equation}
b_{m}^{(i)}=\sum_{n}A_{m,n}^{(i)}x_{n}.\label{eq:op_coefficients}
\end{equation}
$A^{(i)}$ can be approximated as
\[
A^{(i)}=T^{(i)}C\quad\mbox{or}\quad A^{(i)}=CT^{(i)},
\]
where $C$ represents the effect of the point-spread function as a
convolution, while $T^{(i)}$ represents geometric transformation
and down-sampling. While not explicitly shown in the above description,
we choose to express the geometric transformation itself as a linear
transformation, $\mathbf{p}'=H\mathbf{p}$, where $\mathbf{p}$ is
a homogeneous pixel coordinate. Non-linear transformation models are
just as viable, but for simplicity we choose to estimate a homeography
$H_{i,0}$ during registration. That homeography transforms the low-resolution
image $i$ onto the reference frame $i=0$ (which is typically the
first image in the sequence, but can be chosen arbitrarily).

The application order of the operators $T^{(i)}$ and $C$ is not
arbitrary, and either choice presents difficulties. Importantly, $T^{(i)}$
transforms a high-resolution image into a low-resolution image. Therefore,
if the convolution $C$ is applied first, forshortening due to the
geometric transformation may lead to certain areas of the scene being
more densely sampled than others. If $T$ is applied first, i.e. if
the convolution operator acts on the resulting low-resolution image,
some samples in the high-resolution image may not be taken into consideration
at all.

Capel in \citep{Capel2001} addresses this problem by designing the
matrix $A$ as a convolution kernel (representing the camera point-spread
function), transformed by the known geometric transformation $H$
to modify its shape and convolution path. One could imagine this warped
convolution ``fetching'' all high-resolution pixels that contribute
to a specific low-resolution pixel. The approach works well, but introduces
some challenges of its own: what, e.g., should the shape and size
of the convolution kernel be? The camera point spread function is
well modelled as a Gaussian kernel, but even so the variance, $\sigma^{2}$,
is unknown. How should the transformed kernel be represented and applied?
Capel models the kernel as a piecewise bilinear surface to allow transformation
and integration. The variance parameter may be estimated by repeatedly
making reconstructions while varying $\sigma^{2}$ until the result
shows little oscillatory behaviour.

Still, we would prefer not to have such free parameters at all, which
leads to the simplified operator, $A^{(i)}=T^{(i)}$, proposed in
the next section.

\selectlanguage{english}%

\section{Interpolation operators}

\selectlanguage{british}%
\begin{figure}
\begin{centering}
\subfloat[Input image (one of ten), upscaled using sinc (Lanczos) interpolation.]{\begin{centering}
\includegraphics[width=0.48\textwidth]{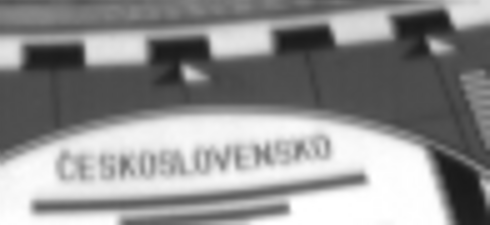}
\par\end{centering}

}~~~\subfloat[Reconstruction at $5\times$ zoom with bilinear interpolation. A high-resolution
reconstruction is made, but the result is oscillatory due to the bilinear
operator's small footprint.]{\begin{centering}
\includegraphics[width=0.48\textwidth]{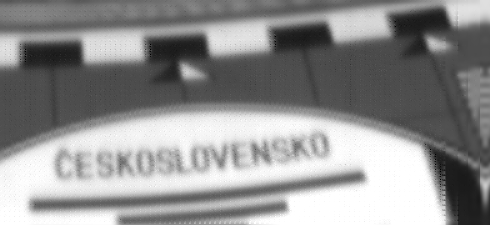}
\par\end{centering}

}~~~
\par\end{centering}

\begin{centering}
\subfloat[Reconstruction at $1.8\times$ zoom with bilinear interpolation. Note
that, while the resolution of this reconstruction is low, the detail
is at least as good as the $5\times$ reconstruction above, but without
oscillations. The small footprint of the bilinear super-resolution
operator is adequate for such low zoom factors.]{\begin{centering}
\includegraphics[width=0.48\textwidth]{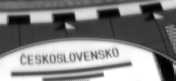}
\par\end{centering}

}~~~\subfloat[Reconstruction at $5\times$ zoom with polygon interpolation. This
operator has a variable size footprint, and is therefore capable of
handling any size zoom factor.]{\begin{centering}
\includegraphics[width=0.48\textwidth]{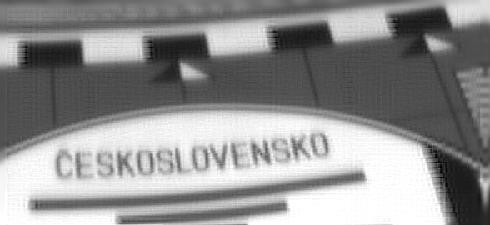}
\par\end{centering}

}
\par\end{centering}

\caption{\label{fig:bilinear_coverage}The effect of the zoom factor on bilinear
and polygon super-resolution operators.}
\end{figure}

By neglecting the effect of the camera point spread function, the
image formation matrix, $A^{(i)}$, can be approximated as the outcome
of the geometric warp and down-sampling,
\[
A^{(i)}=T^{(i)}.
\]
Imagine that $T^{(i)}$ is implemented as bilinear interpolation,
then this simplification introduces the obvious flaw that only 4 high-resolution
pixels are used to calculate the value of any single low-resolution
pixel. In reality, a low-resolution pixel may (and probably will)
depend on more high-resolution pixels; the exact number being determined
by the resolution increase, $z$, and the severity of the transformation,
$H$. If the zoom ratio is chosen conservatively, the approximation
may be sufficient, as illustrated in Figure~\ref{fig:bilinear_coverage}.

Next, we examine the construction of such a \emph{bilinear interpolation}
operator. An improved \emph{polygon-based interpolation} operator
is then proposed to address its deficiencies.

\subsection{Bilinear interpolation}

The bilinear transformation/interpolation operator, $A=T$, has coefficients
appearing on and around the diagonal. The coefficients in \eqref{eq:op_coefficients}
are derived from bilinear interpolation as follows:

Suppose a function is known at four grid coordinates surrounding $(x,y)$,
namely $f_{00}=f(\left\lfloor x\right\rfloor ,\left\lfloor y\right\rfloor )$,
$f_{01}=f(\left\lfloor x\right\rfloor ,\left\lfloor y\right\rfloor +1)$,
$f_{10}=f(\left\lfloor x\right\rfloor +1,\left\lfloor y\right\rfloor )$,
and $f_{11}=f(\left\lfloor x\right\rfloor +1,\left\lfloor y\right\rfloor +1)$.
The value at $f(x,y)$ is not available, but a two-fold linear interpolation
approximates it as
\begin{multline}
f(x,y)\approx\left[\begin{array}{cc}
1-u & u\end{array}\right]\left[\begin{array}{cc}
f_{00} & f_{01}\\
f_{10} & f_{11}
\end{array}\right]\left[\begin{array}{c}
1-t\\
t
\end{array}\right]\\
=f_{00}(1-u)(1-t)+f_{01}(t)(1-u)+f_{10}(u)(1-t)+f_{11}(u)(t)\label{eq:bilin_coeff_derive}
\end{multline}
with $u=x-\left\lfloor x\right\rfloor $ and $t=y-\left\lfloor y\right\rfloor $.
This is known as bilinear interpolation (even though the successive
combination of two linear operators is no longer linear). If all known
grid-values of $f(x,y)$ are placed in a vector, $\mathbf{x}$, then
\begin{equation}
f(x,y)\approx\mathbf{a}^{T}\mathbf{x},\label{eq:coeff_vec_mat}
\end{equation}
where $\mathbf{a}$ is a sparse vector of interpolation coefficients
given by \eqref{eq:bilin_coeff_derive} ($\mathbf{a}$ has mostly
zero entries, except where elements in $\mathbf{x}$ correspond to
$f_{00},$ $f_{01}$, $f_{10}$, or $f_{11}$). When interpolating
for several coordinate pairs $(x_{i},y_{i})$ simultaneously, \eqref{eq:coeff_vec_mat}
becomes
\begin{equation}
\mathbf{b}=A\mathbf{x},\quad\mathrm{with}\,\mathbf{b}=\left[\begin{array}{c}
f(x_{0},y_{0})\\
f(x_{1},y_{1})\\
\vdots\\
f(x_{N-1},y_{N-1})
\end{array}\right].\label{eq:axb_interp_map}
\end{equation}

\begin{figure}
\centering{}\subfloat[\label{fig:bilinear_pixel_warp}With bilinear interpolation, the centre
of the transformed pixel (big dot) is used to find the nearest surrounding
pixel centres (small dots). Only those high-resolution pixels are
used to compute the value of the low-resolution pixel, weighted according
to their distance from the center of the transformed pixel.]{\begin{centering}
\includegraphics[width=0.4\textwidth]{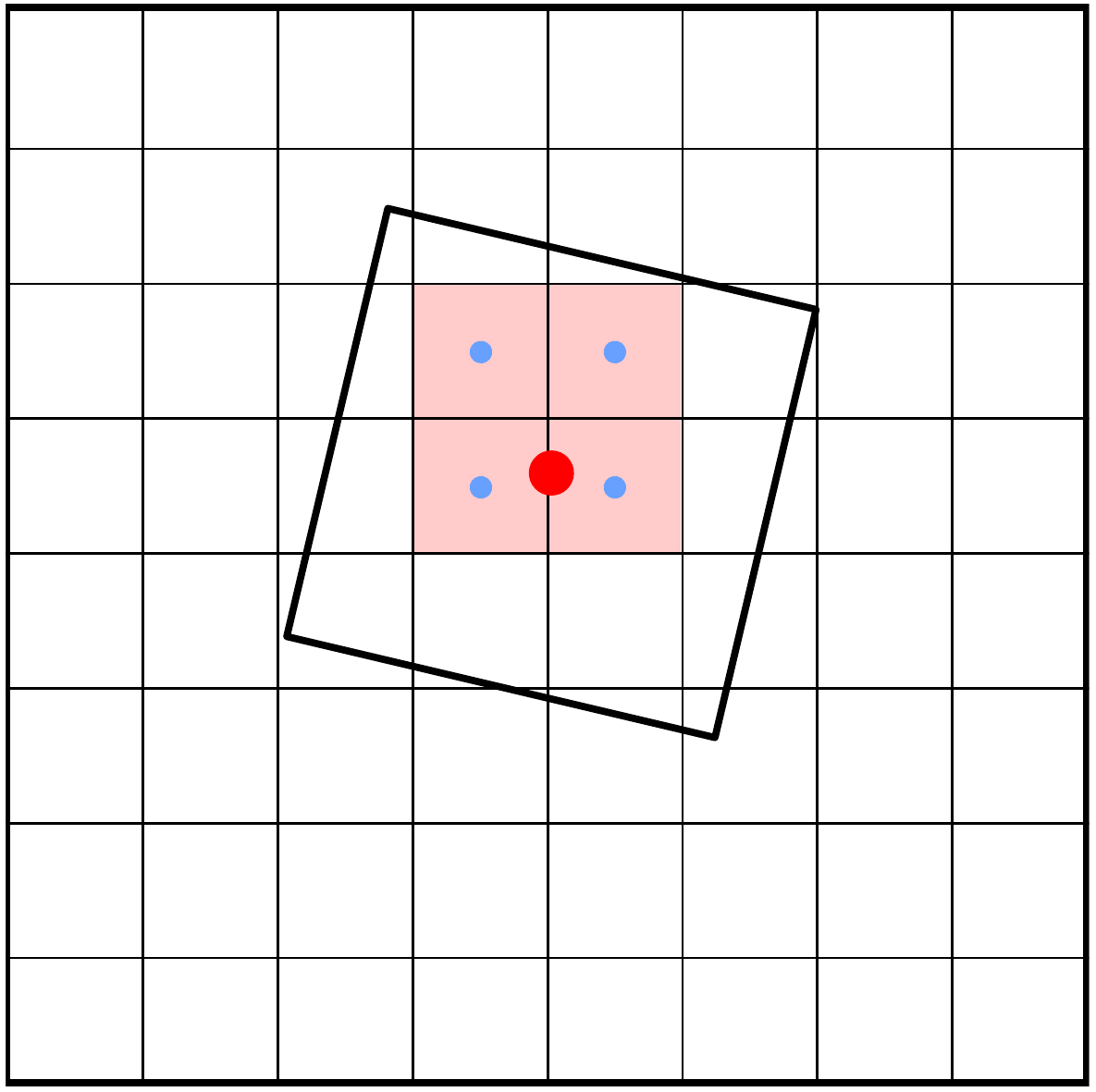}
\par\end{centering}

}~~~~\subfloat[\label{fig:Polygon-interpolation-footprint}With polygon-based interpolation,
all high-resolution pixels touching the transformed pixel are used
in computing the value of the low-resolution pixel, according to the
amount of overlap with the transformed pixel.]{\begin{centering}
\includegraphics[width=0.4\textwidth]{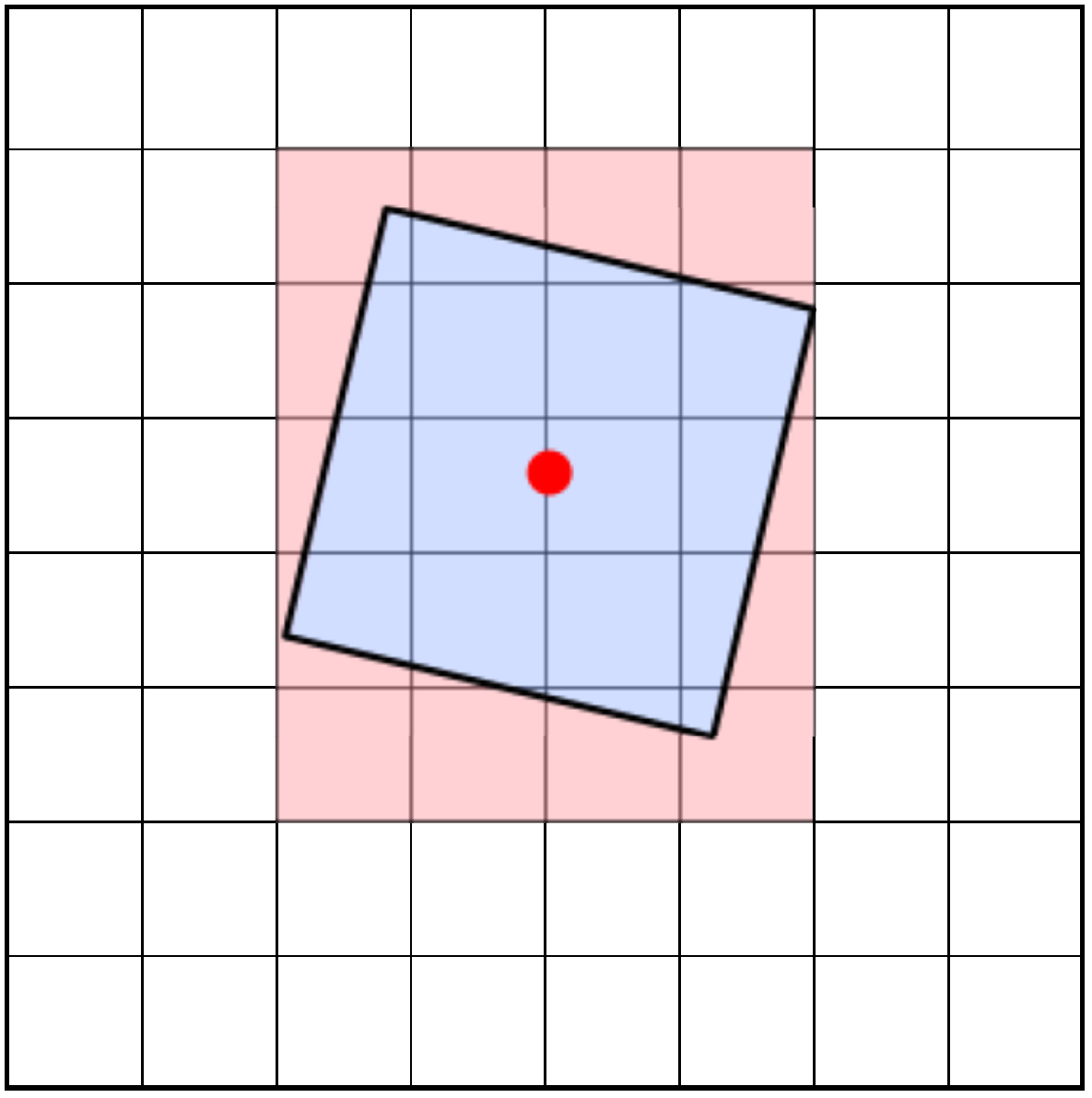}
\par\end{centering}

}\caption{\selectlanguage{english}%
A comparison between the footprints of bilinear and polygon-based
interpolation. \foreignlanguage{british}{Each figure shows a low-resolution
pixel, transformed onto the high-resolution image.}\selectlanguage{british}%
}
\end{figure}

\begin{figure}
\selectlanguage{british}%
\centering{}\subfloat[High-resolution input image. The vector representation, $\mathbf{x}$,
is obtained by unpacking the values in lexicographic order.]{\centering{}\includegraphics[scale=0.3]{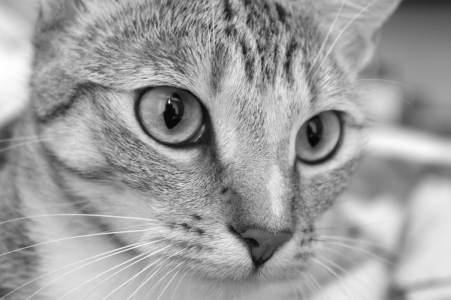}}~~~~\subfloat[\label{fig:Ax_illustration}The matrix-vector product, $A\mathbf{x}$,
for the bilinear operator, reshaped to form an image. ]{\begin{centering}
\includegraphics[scale=0.3]{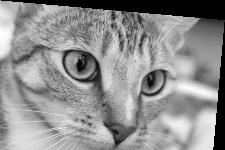}
\par\end{centering}

}~~~~\subfloat[The matrix-vector product, $A\mathbf{x}$, for the polygon interpolation
operator.]{\begin{centering}
\includegraphics[scale=0.3]{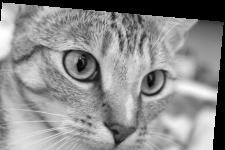}
\par\end{centering}

}\foreignlanguage{english}{\caption{The effect of interpolation operators.}
}\selectlanguage{english}%
\end{figure}

For super-resolution reconstruction, we need $A^{(i)}$ to warp and
down-scale the high-resolution image, $\mathbf{x}$, to produce a
given low-resolution image, $\mathbf{b}^{(i)}$. Recall that, after
registration, the homeography $H_{i,0}$ that warps low-resolution
image $i$ onto low-resolution reference image ($i=0$) is known.
The homography that warps the \emph{high-resolution} image to the
$i$-th low-resolution image then becomes
\[
M_{i}=\left(H_{i,0}\right)^{-1}S\quad\mbox{with}\quad S=\left[\begin{array}{ccc}
1/z & 0 & 0\\
0 & 1/z & 0\\
0 & 0 & 1
\end{array}\right].
\]
Rewriting \eqref{eq:axb_interp_map} as
\[
\mathbf{b}^{(i)}=A^{(i)}\mathbf{x},\quad\mathbf{b}^{(i)}=\left[\begin{array}{c}
f(M_{i}^{-1}\mathbf{c}_{0})\\
f(M_{i}^{-1}\mathbf{c}_{1})\\
\vdots\\
f(M_{i}^{-1}\mathbf{c}_{N-1})
\end{array}\right]
\]
where $\mathbf{c}_{j}$, $j=0,\ldots,N-1$ represents all coordinates
in a low-resolution frame. With the coordinates $M_{i}^{-1}\mathbf{c}_{j}$
known, the matrix $A^{(i)}$ can be constructed. Figure \ref{fig:Ax_illustration}
illustrates the effect of $A$ on $\mathbf{x}$.

\begin{figure}
\centering{}\subfloat[Sparsity structure of the bilinear interpolation operator $A$. Note
that no row contains more than 4 non-zero values.]{\selectlanguage{english}%
\begin{centering}
\includegraphics[width=0.4\textwidth]{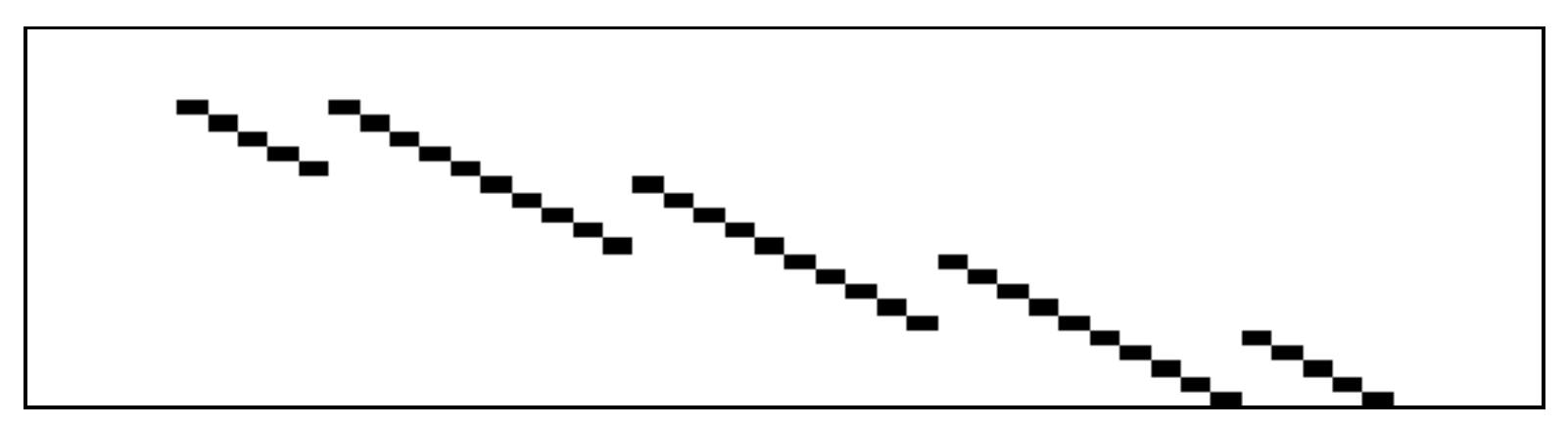}
\par\end{centering}

\selectlanguage{british}%
}~~~~\subfloat[Sparsity structure of the polygon interpolation operator $A$. The
first rows are empty, since the corresponding pixels fall outside
of the input image. The rows contain more coefficients than in the
case of the bilinear operator, indicating a larger footprint.]{\selectlanguage{english}%
\begin{centering}
\includegraphics[width=0.4\textwidth]{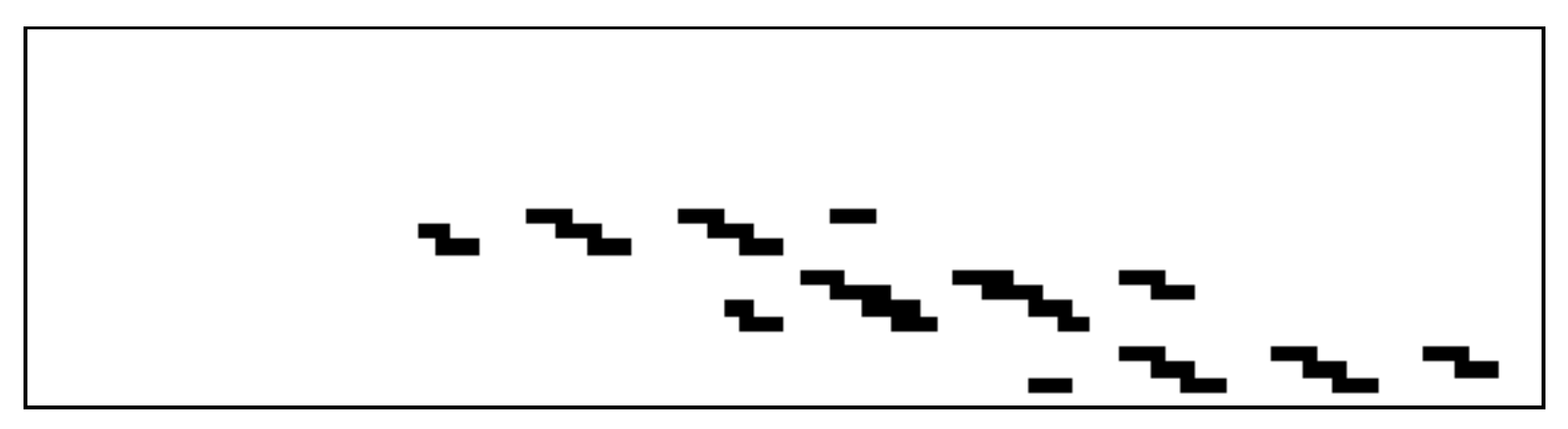}
\par\end{centering}

\selectlanguage{british}%
}~~~\caption{\label{fig:operator_sparsity}Comparison of the structures of the
bilinear and polygon image formation matrices. In both cases, $A$
was constructed to rotate by $5^{\circ}$ and to downsample by 2.
Only the first few rows of each operator is shown.}
\end{figure}

\selectlanguage{english}%

\selectlanguage{british}%

\subsection{Polygon-based interpolation}

In \citep{VanDerWalt2007}, a polygon intersection scheme, subsequently
found to be similar to NASA's Drizzle%
\footnote{\url{http://stsdas.stsci.edu/multidrizzle}%
} \citep{Fruchter2002}, is presented for image fusion. Both methods
rely on intersecting quadrilaterals (four-cornered polygons that represent
pixels) to determine pixel weights, but the formulation as a linear
operator---as discussed in this paper---proves to be fundamental in
its application to super-resolution imaging.

As in the previous section, we want to express a low-resolution output
image as $\mathbf{b}=A\mathbf{x}$. Each pixel value, $b_{m}$, depends
on a number of pixels from the high-resolution image, $\mathbf{x}$,
weighted by the coefficients in row $m$ of the operator $A$. The
motivation behind the polygon interpolation operator is as follows:

A camera sensor is a grid of photo-sensitive cells (think of them
as photon buckets, each representing a pixel \citep{Healey1994}).
Due to micro-lenses, the gaps between cells are negligible. During
imaging, the sensor irradiance is integrated over each cell for the
duration of exposure, after which the values are read out as a matrix.
Now, imagine two sensors, one with large cells (low-resolution) and
the other with small cells (high-resolution), rotated relative to
one another. How are the cell values for the different sensors related?
Our proposed solution is to measure the overlap between the larger
and smaller cells, as shown in Figure~\ref{fig:Polygon-interpolation-footprint}.
The value of a (large) low-resolution cell is set to a weighted sum
of all (small) high-resolution cells, where the weights depend on
their overlap. 

\begin{algorithm}
\medskip{}
For each low-resolution pixel, $b_{m}$:
\begin{enumerate}
\item Create a quadrilateral (four-node polygon) from the corner-points
of $b_{m}$. For example, the pixel at $(0,0)$ would correspond to
the polygon with nodes
\begin{eqnarray*}
\mathbf{x}_{L}^{m} & = & (-0.5,0.5,0.5,-0.5)\\
\mathbf{y}_{L}^{m} & = & (-0.5,-0.5,0.5,0.5).
\end{eqnarray*}
The subscript $L$ indicates ``low-resolution'' and the super-script
is the pixel number.
\item Transform the polygon to the high-resolution frame, using the transformation
matrix $M^{-1}$ given in the previous section. The new corner coordinates
are $\mathbf{\hat{x}}_{L}^{m}$, $\hat{\mathbf{y}}_{L}^{m}$. If any
of the coordinates fall outside the high-resolution image, break this
loop and continue to the next low resolution pixel (there may be other
ways to handle boundary problems, but this is simple and works well).
\item Determine the bounding box of the newly formed polygon:
\begin{eqnarray*}
\mathbf{x}_{BB} & = & \left(\left\lfloor \mbox{\ensuremath{\min}}\mathbf{x}'_{L}\right\rfloor ,\left\lceil \mbox{\ensuremath{\max}}\mathbf{x}'_{L}\right\rceil ,\left\lceil \mbox{\ensuremath{\max}}\mathbf{x}'_{L}\right\rceil ,\left\lfloor \mbox{\ensuremath{\min}}\mathbf{x}'_{L}\right\rfloor \right)\\
\mathbf{y}_{BB} & = & \left(\left\lfloor \mbox{\ensuremath{\min}}\mathbf{y}'_{L}\right\rfloor ,\left\lfloor \mbox{\ensuremath{\min}}\mathbf{y}'_{L}\right\rfloor ,\left\lceil \mbox{\ensuremath{\max}}\mathbf{y}'_{L}\right\rceil ,\left\lceil \mbox{\ensuremath{\max}}\mathbf{y}'_{L}\right\rceil \right)
\end{eqnarray*}

\item For each high-resolution pixel inside the bounding box:

\begin{enumerate}
\item Assign the pixel number $n=iN+j$ where $(i,j)$ is the grid position
of the high-resolution pixel and $N$ is the total number of columns
in the high-resolution frame. 
\item Create a quadrilateral from the corner-points of the high-resolution
pixel with vertices $\mathbf{x}_{H}^{n}$ and $\mathbf{y}_{H}^{n}$.
\item Measure the area of overlap between the polygons $(\mathbf{x}_{L}^{m},\mathbf{y}_{L}^{m})$
and $(\mathbf{x}_{H}^{n},\mathbf{y}_{H}^{n})$, and assign the value
to $A_{m,n}$.
\end{enumerate}
\item Divide each row $A_{m,*}$ by its sum so that the weights add to one.
\end{enumerate}
\caption[Coefficients of the polygon interpolation operator.]{\label{alg:polygon_coefficients}Calculating the coefficients of
the polygon interpolation operator $A$. Also see Figure~\ref{fig:Polygon-interpolation-footprint}.}
\end{algorithm}
Algorithm~\ref{alg:polygon_coefficients} outlines the calculation
of the coefficients in $A$. The operator is parameter free, and has
a variable size footprint that covers all relevant high-resolution
pixels (see Figure \ref{fig:operator_sparsity}). Furthermore, it
has less of a smoothing effect than the bilinear interpolator.

Computing the coefficients of $A$ is more expensive for polygon interpolation
than for bilinear interpolation, due to the many polygon area intersection
calculations (called ``clipping'' operations) involved. However,
two conditions improve execution time when clipping: polygons from
the high-resolution image are always aligned with the grid (horizontal
and vertical boundaries), and all polygons involved are convex.

The first observation is particularly important, since it allows the
use of algorithms that clip polygons to a ``viewport'' (typically
used to determine which part of a polygon falls inside the screen).
The second means that a simpler class of algorithm can be employed.
We use the Liang-Barsky algorithm \citep{Liang1983}, which is specifically
optimized for rapidly clipping convex polygons against a viewport.
Another applicable approach is that by Maillot \citep{Maillot1992}.

The area of the resulting non-self-intersecting clipped polygon is
easily determined as given by \citep{Bourke}: 
\[
a=\frac{1}{2}\sum_{i=0}^{N-1}(x_{i}y_{i+1}-x_{i+1}y_{i}),
\]
where $\mathbf{x}$ and $\mathbf{y}$ are the polygon vertices in
either clock-wise or anti-clock-wise order. The clipping of the entire
collection of pixels can easily be parallellized. Also, if only the
result of the operator, $A\mathbf{x}$, is required, it can be rapidly
rendered via a Graphical Processing Unit without explicitly calculating
any polygon intersections manually (pixels are simply drawn as polygons
and transformed; the GPU takes care of clipping in its fixed pipeline).
Figure~\ref{fig:Ax_illustration} illustrates the effect of $A$
on a vector $\mathbf{x}$.

Note that this interpolation model is only applicable \emph{before}
Bayer demosaicking (a process which destroys much of the super-resolution
information in any case).

\begin{figure}
\begin{centering}
\subfloat[Example input frame (one of thirty), upscaled by a factor of 4 using
sinc (Lanczos) interpolation.]{\includegraphics[width=0.8\textwidth]{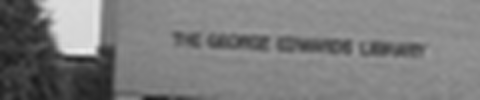}

}
\par\end{centering}

\begin{centering}
\subfloat[All input frames, upsampled and averaged (stacked), after photometric
registration.]{\includegraphics[width=0.8\textwidth]{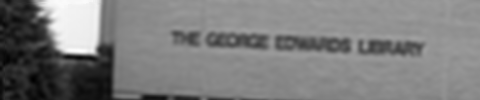}

}
\par\end{centering}

\begin{centering}
\subfloat[Super-resolution result after photometric registration, using polygon
interpolation, zoom factor $z=4$, and $\lambda=0.05$. Note the appearance
of the slanted brick-face.]{\includegraphics[width=0.8\textwidth]{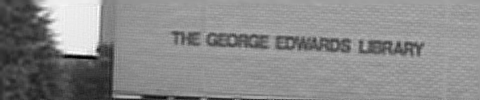}

}
\par\end{centering}

\caption{\label{fig:results}Super-resolution result achieved with polygon-based
interpolation.}
\end{figure}

\selectlanguage{english}%

\section{Results and conclusion}

As test data, we use 30 images and their corresponding homographic
transformations to a chosen reference, obtained from a hand-held video
of a library wall (see acknowledgements). We solve the regularized
least-squares system in \eqref{eq:solve_delta} using conjugate gradients
(L-BFGS \foreignlanguage{british}{\citep{Zhu1997}} and LSQR \foreignlanguage{british}{\citep{Paige1982,Paige1982a}}
give comparable results). The prior $\mathbf{x}_{0}$ is chosen as
the average of all low-resolution images, up-scaled. The high-resolution
estimate $\mathbf{x}$ is then obtained from \eqref{eq:prior_plus_delta}. 

\selectlanguage{british}%
The results shown in Figures \ref{fig:bilinear_coverage} and \ref{fig:results}
prove the viability of the proposed method. \foreignlanguage{english}{Software
that implements the entire super-resolution reconstruction is provided
under an open source license at \url{http://mentat.za.net/supreme}.}

\selectlanguage{english}%
We've shown that, in the context of super-resolution imaging, an interpolation
operator based on polygon intersection\foreignlanguage{british}{ holds
several advantages: it models the underlying sensor physics, it is
easy to compute, has a variable footprint that automatically adjusts
to the underlying transformation and can easily be expressed as a
linear operator. The method can also be extended to model pixels as
higher order polygons, thereby allowing for non-linear image transformations
between input images (such as those introduced by lens distortions).}

More detail on the various aspects of super-resolution imaging presented
here is given in \citep{vdwalt2010}.

\subsubsection*{Acknowledgements}

The library data sequence is by Barbara Levienaise-Obadia, University
of Surrey. \foreignlanguage{british}{Tomas Pajdla and Daniel Martinec,
CMP, Prague provided the ``text'' dataset.} Both were collected
by David Capel and is available for download from \url{http://www.robots.ox.ac. uk/~vgg/data}.

\subsubsection*{References}

\renewcommand{\bibsection}{}

\bibliographystyle{plain}
\bibliography{vdwalt_herbst_polygon_sr,thesis}

\end{document}